\documentclass{article} %
\usepackage{iclr2027_conference,times}

\newenvironment{proofcustom}[1][Proof]
  {\par\noindent\textbf{#1.}\ }
  {\hfill$\square$\par\medskip}
\usepackage{hyperref}
\usepackage{url}
\usepackage{graphicx}

\newtheorem{lemma}{Lemma}

\newtheorem{theorem}{Theorem}
\newtheorem{corollary
}{Corollary
}
\usepackage{booktabs}       %
\usepackage{amsfonts}       %
\usepackage{nicefrac}       %
\usepackage{microtype}      %
\usepackage{comment}
\usepackage{amssymb, amsmath, latexsym}
\usepackage{url}

\usepackage{algorithm}
\usepackage{algorithmic}

\usepackage{tabularx}
\usepackage{paralist}
\usepackage{mathtools}

\usepackage{bbm} %
\usepackage{wrapfig}
\usepackage{makecell}
\usepackage{multirow}
\usepackage{booktabs}

\usepackage{nicefrac}       %

\usepackage{boxhandler}
\usepackage[flushleft]{threeparttable} %

\usepackage{caption}
\usepackage{multirow}
\usepackage{colortbl}
\definecolor{bgcolor}{rgb}{0.8,1,1}
\definecolor{bgcolor2}{rgb}{0.8,1,0.8}
\definecolor{niceblue}{rgb}{0.0,0.19,0.56}

\usepackage{hyperref}
\hypersetup{colorlinks,linkcolor={blue},citecolor={niceblue},urlcolor={blue}}

\usepackage{pifont}
\definecolor{PineGreen}{RGB}{0,110,51}
\definecolor{BrickRed}{RGB}{143,20,2}

\usepackage{tikz-cd} %

\DeclareMathOperator*{\argmin}{arg\,min}

\newcommand{\R}{\mathbb{R}}

\def\<#1,#2>{\left\langle #1,#2\right\rangle}

\newcolumntype{Y}{>{\centering\arraybackslash}X}

\usepackage{xspace}

\newcommand{\mix}{\mathrm{mix}}
\newcommand{\UM}{\mathrm{UM}}
\newcommand{\IDU}{\mathrm{IDU}}

\usepackage[colorinlistoftodos,bordercolor=orange,backgroundcolor=orange!20,linecolor=orange,textsize=scriptsize]{todonotes}

\newcommand{\EE}{\mathbb{E}}

\usepackage{hyperref}
\graphicspath{{plots/}}

\usepackage{makecell}

\usepackage{accents}
\newlength{\dhatheight}

\usepackage{pgfplotstable} %
\usetikzlibrary{automata, positioning, arrows, shapes, fit, calc, intersections}
\usepgfplotslibrary{statistics}

\def\la{\langle}
\def\ra{\rangle}
\definecolor{revisionpurple}{RGB}{128,0,128}
\definecolor{revisionforestgreen}{RGB}{34,139,34}
\title{Data Unlearning via Inverse Distillation}

\author{
\parbox[t]{0.45\textwidth}{\normalfont\raggedright
\textbf{Aleksei Leonov} \\
AI Foundation lab, Moscow, Russia \\
MIRAI, Moscow, Russia \\
\texttt{leonovich1999@gmail.com}
}
\And
\parbox[t]{0.45\textwidth}{\normalfont\raggedright
\textbf{Nikita Kornilov} \\
Applied AI Institute, Moscow, Russia \\
MIRAI, Moscow, Russia \\
BRAIn Lab, Moscow, Russia \\
\texttt{jhomanik14@gmail.com}
}
\AND
\parbox[t]{0.45\textwidth}{\normalfont\raggedright
\textbf{Zhang Zhenhe} \\
AI Foundation lab, Moscow, Russia
}
\And
\parbox[t]{0.45\textwidth}{\normalfont\raggedright
\textbf{Evgeny Burnaev} \\
Applied AI Institute, Moscow, Russia \\
AXXX, Moscow, Russia
}
\AND
\parbox[t]{0.45\textwidth}{\normalfont\raggedright
\textbf{Iaroslav Koshelev} \\
AI Foundation lab, Moscow, Russia
}
\And
\parbox[t]{0.45\textwidth}{\normalfont\raggedright
\textbf{Alexander Korotin} \\
Applied AI Institute, Moscow, Russia \\
AXXX, Moscow, Russia \\
\texttt{iamalexkorotin@gmail.com}
}
}

\iclrfinalcopy
\begin{document}

\maketitle
\lhead{}

\begin{abstract}
Multi-step matching models, including flow and diffusion models, produce high-quality outputs but incur substantial inference costs and may reproduce unwanted components of their training datasets. We introduce Inverse Distillation Unlearning (IDU), a unified framework that simultaneously distills a teacher multi-step matching model into an efficient one-step student generator and suppresses outputs corresponding to a designated training subset. We first formulate distillation as a min-max objective over a data distribution and then represent this distribution as a mixture of the forget-set and the generated distributions. This allows us to compare this mixture with the teacher's training distribution and recover only the retained data  at the optimum. Our method requires only a pretrained full-data teacher and data from the forget set, without access to retained training examples, extra feature extractors or classifiers. Extensive experiments on MNIST and CIFAR-10 datasets under flow-matching and score-based diffusion settings demonstrate that IDU substantially reduces the generation frequency of forgotten classes while preserving generation quality on the retained classes. To the best of our knowledge, IDU is the first unified framework for simultaneous unlearning and distillation in unconditional flow-matching and score-based models.

\end{abstract}

\section{Introduction}

Multi-step diffusion \citep{sohl2015deep, ho2020ddpm, song2021sde}, flow \citep{lipman2023flowmatching, liu2023rectifiedflow}, and other matching models \citep{holderrieth2024generator, gao2025diffusion}, followed by one-step generators \citep{kim2024ctm, zhou2024score, yin2024improved, fransone}, now produce high-quality and diverse samples. Yet their reliance on large, often imperfectly moderated web datasets \citep{schuhmann2022laion} raises privacy, copyright, legal, and ethical concerns \citep{voigt2017eu, goldman2020introduction}: models may memorize and reproduce unwanted training content \citep{carlini2023extracting}.
To avoid such incidents, a critical research direction known as \textit{machine unlearning (MU)} has emerged within the field of trustworthy machine learning \citep{bourtoule2021machine, nguyen2025survey}. At its core, MU aims to remove the influence of specific training data, classes, or semantic concepts from trained generative models, while preserving the same generation quality for the remaining data. The MU literature splits into two main paradigms: \textit{data unlearning and class unlearning}.

\textbf{Data unlearning} \citep{alberti2025siss} addresses scenarios where the model is fine-tuned to completely erase the influence of \emph{particular data} for which no class or prompt anchor is available — such as individual faces or Not-Safe-For-Work (NSFW) images. In this setup, the forget data is typically provided as selected samples.  Ideally, the goal is to obtain a model as if it were trained only on the remaining data, yet without retraining from scratch. Data unlearning naturally appears in unconditional models, but can also be applied to conditional variants when some data should be erased within a given class. 

The key challenge in this setup is that the forget and remaining data are deeply entangled within the model parameters — the forget data cannot be directly prompted during generation. Furthermore, the quality of the remaining data must be preserved, even though its samples may be unavailable. As a result, typical data unlearning algorithms distinguish between remaining and forget data by employing different losses on them \citep{ alberti2025siss, wu2025erasing, Jiang_2025_ICCV}, constrained optimization \citep{khalafi2026unlearning}, variational framework \citep{panda2024variational}, importance sampling \citep{shi2026retrack}, energy functions \citep{simone2025continualflow}, or transport costs \citep{choi2026uotunlearn}. %

Data unlearning algorithms are well-explored for matching models, with both general frameworks and model-specific solutions (e.g., flow matching \citep{simone2025continualflow} or score-based models \citep{Jiang_2025_ICCV}). \emph{Nevertheless, little progress \citep{choi2026uotunlearn} has been made toward effective, easy-to-tune, and model-agnostic approaches for one-step generators.}

 \textbf{Class unlearning} \citep{gandikota2023erasing} aims to remove \textit{all knowledge} pertaining to unwanted classes in the conditional models. More specifically, unlearned models are fine-tuned to output noise or irrelevant data when conditioned on a particular class, such as an unsafe  object category,  while preserving the generative quality for all other classes. In contrast to data unlearning, the forget data in this setup is typically provided as class labels, without any data samples. Concept unlearning extends this idea to text-to-image models, where the goal is to remove entire concepts or styles (e.g.,  artistic style, nudity or  celebrity's likeness) that may be triggered by textual prompts.

Most class unlearning approaches optimize a combination of two losses: a forget loss that swaps the undesirable class with another one, and a remaining loss that preserves quality for other classes. Various techniques have been developed for designing and combining these losses for diffusion and flow models, including steer-away guidance \citep{gandikota2023erasing}, Bayesian continual learning \citep{heng2023selective}, saliency-based weight updates \citep{fan2024salun}, cross-attention editing \citep{gandikota2024unified, lu2024mace}, attention re-steering \citep{zhang2024forget}, pruning \citep{chavhan2024conceptprune}, adversarial strategies \citep{bui2025fantastic}, multi-objective optimization to avoid conflicts between loss gradients \citep{wu2025erasing}, class swapping during distillation \citep{chen2025sfd}, attention regularization  \citep{pmlr-v267-gao25j, fan2026eraseanythingplusplus}, and  unlearning irreversibility techniques \citep{sharma2024unlearningorconceal, liu2025erased, lu2026concepts}.  \textit{However, only a few papers \citep{chen2025sfd} tackle class unlearning in one-step models, motivating further research.}

\subsection{Contributions}

We fill the gap in data unlearning for one-step models and propose our novel {\textbf{Inverse Distillation Unlearning (IDU)}} approach. Our IDU can erase undesired data samples from a pretrained one-step generator or build a new one without them, using only the teacher matching model trained on the original dataset. Our method follows an inverse distillation pipeline and compares the current mixture of the generated and forget data with the teacher's correct one, penalizing the generator for reproducing unwanted samples. Moreover, IDU can work with different matching teachers, such as diffusion or flow models, and requires neither extra training-time feature extractors nor classifiers; its only method-specific trade-off parameter is $\rho \in [0,1)$.

\section{Related Work}
\subsection{Diffusion, flow and matching models}

Denoising diffusion probabilistic model \citep[\textbf{DDPM}]{ho2020ddpm} defines a multi-step forward process, mapping data to Gaussian noise, and learns to reverse it via a trained denoiser. Score-based generative model \citep[\textbf{SGM}]{song2021sde} extends this idea to continuous time and approximates a score function to simulate the reverse SDE. Flow matching \citep[\textbf{FM}]{lipman2023flowmatching} instead learns an ODE drift that interpolates between data and noise, enabling fewer sampling steps via advanced ODE solvers and flexible interpolations. In all of the above cases, a model must approximate an intractable reverse-process function (denoiser, score, drift, etc.). It is done via available unbiased estimates of this function,  conditioned on initial data samples. Since a model is usually matched with these conditional function estimates, we refer to such models collectively as \textit{matching models}.

Formally, a matching model constructs a probability path $p_t$ on the time interval $[0,T]$, transforming the selected data $p_0$ to noise $p_T$. This path $p_t(x_t) = \int_{\R^D} p_t(x_t|x_0) p_0(x_0) dx_0$ is built as a mixture of simple conditional paths $p_t(\cdot|x_0)$ conditioned on samples $x_0 \sim p_0$. Then, the standard universal matching (\textbf{UM}) loss $\mathcal{L}_{\text{UM}} (f, p_0)$ matches a model $f: [0,T] \times \R^D \to \R^D$ with conditional estimates $f^{p_0}_t(\cdot|x_0 )$ at each time $t$ and point $x_t \sim p_t$:
\begin{equation}
    \!\!\mathcal{L}_{\text{UM}}(f, p_0)\!\! := \EE_{t, x_0 \sim p_0, x_t\sim p_t(\cdot| x_0)}  [\| f_t(x_t) - f^{p_0}_t(x_t|x_0) \|^2]. \label{eq: flow loss}
\end{equation}
Here, the notation $\EE_t$ hides the time sampling and loss weighting inherent to the given matching model. For example, denoising models recover unnoised samples $f_t^{p_0}(x_t|x_0) = x_0$, score-based models match the conditional score $f_t^{p_0}(x_t|x_0) = \nabla_{x_t} \ln p_t(x_t|x_0)$, and flow models with the linear interpolation $x_t=(1-t/T)x_0+(t/T)x_T$ match the conditional drift $f_t^{p_0}(x_t|x_0)=(x_T-x_0)/T=(x_t-x_0)/t$ for $t>0$.

\subsection{Distillation and one-step models}
Multi-step sampling makes matching models slower than one-step generators such as VAEs \citep{kingma2013auto} and GANs \citep{goodfellow2014generative}. Beyond sampling acceleration \citep{lu2022dpm, karras2024analyzing}, matching   distillation methods \citep{yin2024one} train a \textit{one-step generator} $G_\theta$ under a
\textit{multi-step teacher} $f^*$: a fake model fits the generated
distribution, while the generator reduces its discrepancy from the teacher.
Despite differences in model type and discrepancy measure \citep{yin2024improved, zhou2024score, gushchin2025inverse}, these methods admit a common inverse-optimization view \citep{kornilov2026universal}. Given $f^* = \argmin_f \mathcal{L}_{\text{UM}} (f, p_0^*)$ trained via UM loss minimization, they recover its data distribution $p_0^*$. To do this, they optimize the following min-max inverse distillation scheme over trainable generated distribution $p_0^\theta$:
\begin{equation}
    \min{_{\theta}} \max{_f}\left\{ \mathcal{L}_{\text{UM}} (f^*, p^\theta_0) - \mathcal{L}_{\text{UM}} (f, p^\theta_0)  \right\} = \min{_{\theta}} \bigl \{ \mathcal{L}_{\text{UM}} (f^*, p_0^\theta) - \min{_f}\{\mathcal{L}_{\text{UM}} (f, p_0^\theta) \} \bigr\}. \label{eq: inv general form}
\end{equation}
The non-negative difference between losses in this scheme measures how well the teacher fits the current data compared to the best possible fake model. When the teacher data is retrieved, i.e., $p^\theta_0 = p_0^*$, the difference becomes 0 and the scheme attains optimum. %

\subsection{Data unlearning}
In data unlearning, we have an original data distribution $p^{*}_0$, a forget-data distribution $p^{F}_0 $ whose influence we would like to remove, and the remaining data $p^{R}_0$. The ideal solution would be a model trained on the remaining data $p^{R}$ via the vanilla loss $\mathcal{L}_{\text{vanilla}}$. However, instead of training from scratch, the early works \citep{golatkar2020eternal, thudi2022unrolling, tang2026sharpness} propose to fine-tune a weighted sum of the forget  and remaining losses with a trade-off factor $\rho\in [0,1)$:
\begin{equation}
    \mathcal{L}_{\text{base}} = \rho \cdot  \mathcal{L}_\text{forget} + (1 - \rho) \cdot \mathcal{L}_\text{remain}, \label{eq: siss}
\end{equation}
where the forget loss $\mathcal{L}_\text{forget}$ keeps the model away from the forget data,  whereas the remaining loss $\mathcal{L}_\text{remain}$ ensures that the quality on the remaining samples 
does not degrade.

For \textit{unlearning in matching models}, the typical losses are
$\mathcal{L}_{\text{vanilla}} (f) = \mathcal{L}_{\text{UM}} (f, p_0^{R})$, $\mathcal{L}_{\text{forget}}(f) = -\mathcal{L}_{\text{UM}}(f, p_0^{F})$, and
$\mathcal{L}_{\text{remain}}(f) = \mathcal{L}_{\text{UM}}(f, p_0^{R})$. Many data unlearning methods modify the forget and remaining losses along with the optimization procedure between them. \textbf{NegGrad}  minimizes only forget loss $\mathcal{L}_\text{forget}(f)$. However, this approach  often leads to instability and catastrophic forgetting, degrading the overall generation quality. \textbf{SA} \citep{heng2023selective} adds a computationally heavy Elastic Weight Consolidation penalty to the base loss -  the quadratic form of divergence from the initial weights, computed using the Fisher Information Matrix on the current generated data.  \textbf{SalUn} \citep{fan2024salun} optimizes the base loss and applies a binary gradient mask, selecting only the most significant parameters for forgetting. This mask is calculated by thresholding the gradient magnitudes of the forget loss.  \textbf{SISS} \citep{alberti2025siss}  employs importance sampling to call the model only once per base loss $\mathcal{L}_{\text{base}}(f)$ calculation, but basically does not change the loss structure.  \textbf{MGSM} \citep{Jiang_2025_ICCV} incorporates more natural score-function orthogonality instead of $\ell_2$-loss for the forget part.  \textbf{EraseDiff} \citep{wu2025erasing} utilizes random-noise matching loss on forget samples and employs constrained optimization to smoothly merge the gradient directions of the losses. \textbf{Retrack} \citep{shi2026retrack} equips the remaining loss $\mathcal{L}_\text{remain}(f)$ with the nearest-neighbor importance weighting from the forget subset. The work \citep{khalafi2026unlearning} optimizes KL divergence between noising processes, also under the constrained optimization formulation.

Other methods use various types of guidance to avoid the forget data.  \textbf{VDU} \citep{panda2024variational} uses a variational inference framework with a plasticity inducer for reducing the likelihood of unwanted data and a stability regularizer for quality preservation. For flow models, \textbf{ContinualFlow} \citep{simone2025continualflow} applies energy-function weighting to the original loss to suppress unwanted data. %

\textit{One-step model unlearning.} 
\textbf{UOT-Unlearn} \citep{choi2026uotunlearn} uses unbalanced optimal transport to shift a pretrained one-step generator away from unwanted samples. Its cost function requires a precomputed feature extractor and hyperparameter  tuning rather than a trained teacher. 
However, the OT framework  has limited scalability and generation diversity, compared with matching models  distillation.
The same work adapts VDU, SalUn, and SA to consistency models \citep{kim2024ctm, geng2025meanflow, fransone}; these adaptations use a classifier to identify forget samples and do not directly extend to non-self-consistent generators.

\subsection{Class unlearning} Class unlearning aims to erase entire unwanted classes from a conditional model while keeping  other classes intact. Unlike specific data samples, whose influence in an unconditional model is difficult to trace, classes in a conditional model can be efficiently prompted and distinguished from one another. For the same reason, the majority of class-forgetting methods are data-free. These features enable a variety of methods for \textit{matching model unlearning} that bind the attributes of the unwanted classes to completely different entities, especially within the attention or inference structure \citep{gandikota2023erasing, gandikota2024unified, lu2024mace, zhang2024forget, pmlr-v267-gao25j, fan2026eraseanythingplusplus}. Such methods are not always adaptable to data unlearning; nevertheless, they often use the same combination of the forget and remaining losses: the forget loss changes the model's behavior on unwanted classes, while the remaining loss preserves it on the others. For example, SA \citep{heng2023selective}, SalUn \citep{fan2024salun}, and EraseDiff \citep{wu2025erasing} can be leveraged for both data and class unlearning tasks.

\textit{One-step model unlearning.} \textbf{SFD} \citep{chen2025sfd} performs class forgetting during conditional inverse distillation \eqref{eq: inv general form} by replacing the teacher score for forgotten classes with a safe-class score in the generator loss; its full generator losses are given in Appendix~\ref{app:sfd_losses}.

\section{\texorpdfstring{{Inverse Distillation Unlearning}}{Inverse Distillation Unlearning}}

\subsection{Method Description}
\paragraph{Setup and preliminaries.} In data unlearning, we are given a forget-data distribution $p^{F}_0$ that we would like to remove from the original distribution $p^{*}_0$, so that only the remaining (or retained) data $p^{R}_0$ is preserved:
\begin{equation}
p^{*}_0 = \pi\, p^{F}_0 + (1 - \pi)\, p^{R}_0,    \label{eq: data split}
\end{equation}
where $\pi \in [0,1)$ is the proportion of the forget data. In our setup, neither the original nor the remaining data is available, but we do have access to a teacher model $f^* = \argmin_f \mathcal{L}_{\mathrm{UM}}(f, p^{*}_0)$ of an arbitrary matching type, trained on the original data. We aim to train a one-step generator $G_\theta: \mathcal{Z} \to \R^D$
with parameters $\theta$ that will eventually reproduce only the remaining
data $p_0^{R}$. The generator  maps the latent distribution $p_\mathcal{Z}$
to a distribution $p_0^\theta$ and can be pretrained or initialized with a one-step teacher inference scheme.

The standard way to distill the data $p^*_0$, stored inside the teacher model $f^*$, into the trainable distribution $p_0$ is to apply the inverse distillation scheme \eqref{eq: inv general form}. This min-max scheme reverses the forward minimization problem for obtaining the teacher from the fixed input data, i.e., it retrieves the data from which the fixed teacher was obtained:
\begin{equation}
    \min{_{p_0}} \max{_f}\left\{ \mathcal{L}_{\text{UM}} (f^*, p_0) - \mathcal{L}_{\text{UM}} (f, p_0)  \right\} \sim \min{_{p_0}} \bigl \{ \underset{\geq 0 }{\underbrace{\mathcal{L}_{\text{UM}} (f^*, p_0) - \min{_f}\{\mathcal{L}_{\text{UM}} (f, p_0)}} \} \bigr\}. \label{eq: inv general form 2}
\end{equation}
The non-negative difference between losses in this scheme measures how well the teacher fits the current data compared to the best possible fake model. For the teacher data $p_0 = p_0^*$, the optimum is attained with the zero difference.
\paragraph{Our approach.} We build our {\textbf{Inverse Distillation Unlearning (IDU)}} method as follows: we run the inverse distillation scheme \eqref{eq: inv general form 2} but parametrize the optimized distribution $p_0$ as the mixed data $p_0 = \rho\, p^{F}_0 + (1 - \rho)\, p^{\theta}_0 =: p^\mix_0$, similar to the data mix \eqref{eq: data split} with the proportion $\rho \in [0,1)$, where we substitute the desired remaining data $p^R_0$ with the generated one $p_0^\theta$.  Thus, the optimal generator with the right proportion $\rho = \pi$ has to learn only the remaining data in order to recover the full teacher distribution, since {\color{black}the forget component is already accounted for by the forget samples}. The following theorem formalizes the generator's forgetting property. The proof is given in Appendix~\ref{app:idu_proof}.
\begin{theorem}[\textbf{IDU's forgetting property}]\label{thm:idu_forgetting}
    Optimization of IDU loss \eqref{eq: 4 losses} with $\rho = \pi$ retrieves only the remaining data, i.e., the optimal generator parameters $\theta_{\text{opt}}$ yield $p^{\theta_{\text{opt}}}_0 = p_0^R$.
\end{theorem}
More specifically, we optimize the following \textit{min-max IDU objective}  $\mathcal{L}_{\text{IDU}}(f, p_0^\theta) $ over generator parameters $\theta$ and fake model $f$:
\begin{eqnarray}
\mathcal{L}_{\text{IDU}}(f, p_0^\theta)  &:=& \mathcal{L}_{\text{UM}} (f^*, p_0^\mix) - \mathcal{L}_{\text{UM}} (f, p_0^\mix) \notag \\
&=& \mathcal{L}_{\text{UM}} (f^*, \rho\, p^{F}_0 + (1 - \rho)\, p^{\theta}_0) - \mathcal{L}_{\text{UM}} (f, \rho\, p^{F}_0 + (1 - \rho)\, p^{\theta}_0) \notag \\
    &=& \rho \cdot [\mathcal{L}_{\text{UM}}(f^*, p^F_0) - \mathcal{L}_{\text{UM}}(f, p^F_0)] + (1 - \rho) \cdot [\mathcal{L}_{\text{UM}}(f^*, p^\theta_0) - \mathcal{L}_{\text{UM}}(f, p^\theta_0)], \label{eq: 4 losses}
\end{eqnarray}
where the second equality holds because the UM losses \eqref{eq: flow loss} are linear in the input data, which appears only inside a mathematical expectation. 

\subsection{Practical details}

\paragraph{Optimization procedure.} We alternate between two steps to optimize our min-max IDU loss \eqref{eq: 4 losses}:

1) {First, we update the fake model $f$ via the fake loss $\mathcal{L}_{\text{IDU-fake}}(f)$:}
\begin{align}
     &\mathcal{L}_{\text{IDU-fake}}(f) = \rho \cdot \mathcal{L}_{\text{UM}}(f, p^F_0) + (1-\rho) \cdot \mathcal{L}_{\text{UM}}(f, p^\theta_0)  \notag \\
    &=  \EE_{\substack{t, x_0^\theta \sim p^\theta_0, x_t^\theta \sim p^\theta_t(\cdot| x_0^\theta) \\x_0^F \sim p^F_0, x_t^F \sim p^F_t(\cdot| x_0^F) }}  [\rho\| f_t(x^F_t) - f^{F}_t(x_t^F|x_0^F) \|^2 + (1-\rho) \| f_t(x_t^\theta) - f^{\theta}_t(x_t^\theta|x_0^\theta) \|^2 ],  \label{eq: idu fake loss} 
\end{align}
where $p^\theta_t(\cdot| x_0^\theta)$ and $p^F_t(\cdot| x_0^F)$ are the conditional forward noising processes built on the generated and forget data with the corresponding conditional estimates $f^{\theta}_t(x_t^\theta|x_0^\theta)$ and $f^{F}_t(x_t^F|x_0^F)$.

2) Next, we update the generator parameters $\theta$ with a fixed fake model $f$ via the generator loss $\mathcal{L}_{\text{IDU-gen}}(p_0^\theta)$; however, instead of the default loss $\mathcal{L}_{\text{IDU-gen}}(p_0^\theta) = \mathcal{L}_{\text{UM}}(f^*, p_0^\theta)
- \mathcal{L}_{\text{UM}}(f, p_0^\theta),$ we use its modified version, proposed in the Score identity Distillation (SiD) framework \citep{zhou2024score}:
\begin{align} 
      \mathcal{L}_{\text{IDU-gen}}(p_0^\theta) 
&=2 \EE_{\substack{t, x_0^\theta \sim p^\theta_0, \\ x^\theta_t \sim p^\theta_t (\cdot| x^\theta_0)}}  \!\![  
     \la f^*_t(x^\theta_t) - f_t(x^\theta_t), f^*_t(x^\theta_t)  - f_t^{\theta}(x^\theta_t|x^\theta_0)\ra -     \alpha_{\text{SiD}}\|f^*_t(x^\theta_t)  - f_t(x^\theta_t)\|^2], \notag%
\end{align}
where we heuristically scale the second term by exactly $\alpha_{\text{SiD}}$ times. This scale factor $\alpha_{\text{SiD}}$ is usually taken from the range $\alpha_{\text{SiD}} \in [0.5, 1.2]$. For example, in case $\alpha_{\text{SiD}} = 0.5$, we end up with the default theoretical loss, whereas greater factor values can yield better performance in practice. Nevertheless, the optimal value depends strongly on the  matching model type and neural network architecture.
\begin{figure}[!htbp]
    \centering
    \includegraphics[width=\linewidth]{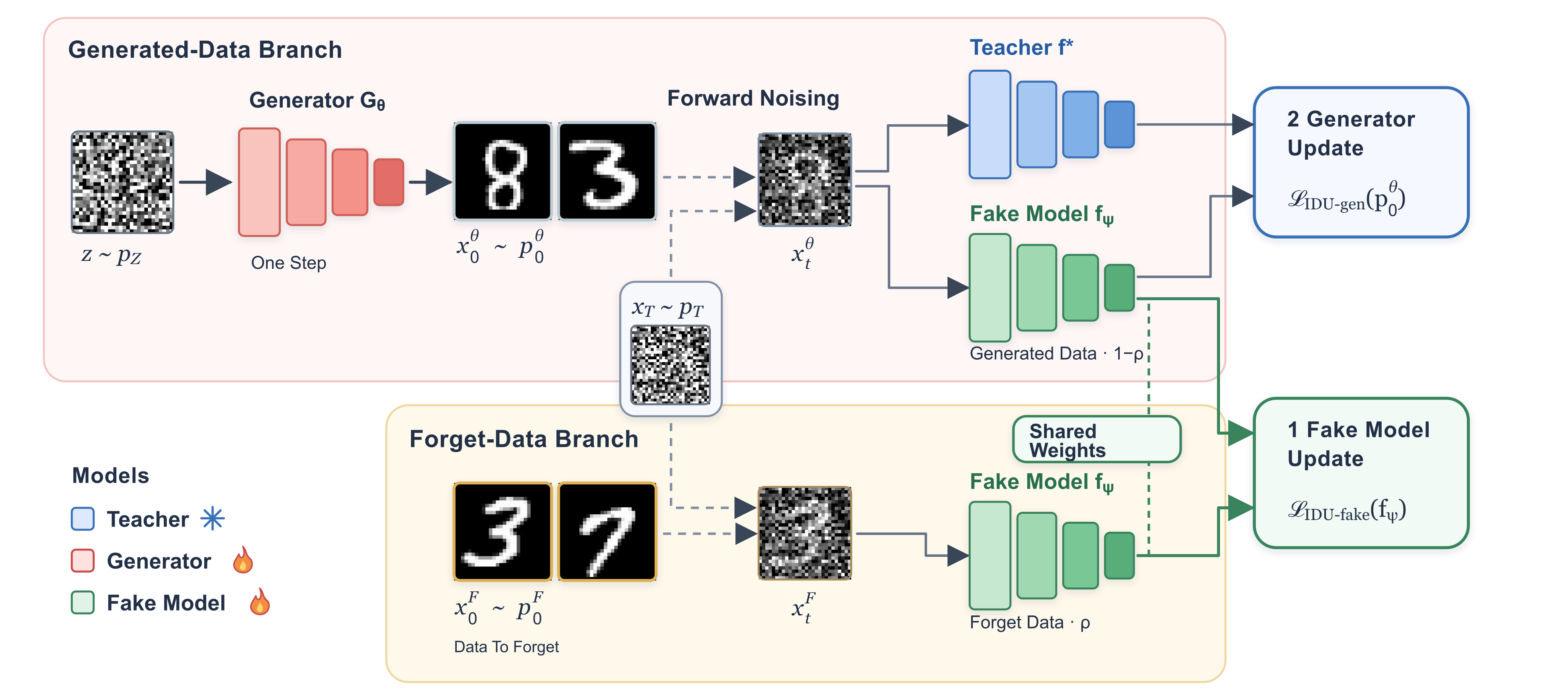}
    \caption{\textbf{Pipeline of our IDU framework}. Forget and generated samples form two forward-noising branches. The framework alternates between two steps:
    first, update the fake model on both branches with weights $\rho$ and $1-\rho$ to detect generations similar to the forget samples, then update the generator to suppress these generations using the frozen teacher and the fake model.}
    \label{fig:idu_pipeline}
\end{figure}

\paragraph{Algorithm pseudocode.} {In Algorithm \ref{alg: IDU}, we provide IDU pseudocode for the flow-matching setting. The general IDU training pipeline is illustrated in Figure~\ref{fig:idu_pipeline}.}

\begin{algorithm}[ht!]
\caption{{Inverse Distillation Unlearning}}
\label{alg: IDU}   
\begin{algorithmic}[1]
\REQUIRE {teacher model $f^*$, generator $G_\theta$ (pretrained or initialized by the teacher), fake model $f_\psi$, forget data $p_0^F$, forgetting scale {$\rho \in [0,1)$}, SiD scale $\alpha_{\text{SiD}} \in [0.5, 1.2]$, number of iterations $N$, batch size $B$, latent distribution $p_\mathcal{Z}$, noise distribution $p_T$.}
\FOR{$n=0,\ldots, N-1$}
\STATE Sample  generated and noise batches $\{x^\theta_{0,i} = G_\theta(z_i)\}_{i=1}^B$, $z_i \sim p_{\mathcal{Z}}$ and $\{x_{T,i}\}_{i=1}^B \sim p_T$;

\STATE Sample times  $\{t_i\}_{i=1}^B$ and noised samples  $\{x^\theta_{t_i, i}\}_{i=1}^B $ according to the model type;

\textcolor{gray}{For flow models: $x^\theta_{t_i, i} = (1-t_i/T)\cdot x^\theta_{0,i} + t_i/T \cdot x_{T,i}$;}

\STATE Sample forget data batch $\{ x^F_{0,i}\}_{i=1}^B \sim p_0^F$ and noised forget samples  $\{ x^F_{t_i,i}\}_{i=1}^B$;

\textcolor{gray}{For flow models: $x^F_{t_i, i} = (1-t_i/T) \cdot x^F_{0,i} + t_i/T \cdot x_{T,i}$;}
\STATE Compute both matching targets  $s^{\theta}_{t_i, i} = f_{t_i}^{\theta}(x^\theta_{t_i, i}|x^\theta_{0, i})$ and $s^{F}_{t_i, i} = f_{t_i}^{F}(x^F_{t_i, i}|x^F_{0, i})$;

\textcolor{gray}{For flow models:  $s^{\theta}_{t_i, i} =   (x_{T,i} - x^\theta_{0,i})/T$ and $s^{F}_{t_i, i} =   (x_{T,i} - x^F_{0,i})/T$;}

\STATE {Update fake model parameters $\psi$ via loss:}
$$  \frac1B\sum\limits^B_{i=1}\left[\rho\| f_{\psi, t_i}(x^F_{t_i, i}) - s^{F}_{t_i, i}\|^2+(1-\rho)\| f_{\psi, t_i}(x^\theta_{t_i, i})-  s^{\theta}_{t_i, i} \|^2\right];$$

\STATE Update generator parameters $\theta$ via loss:
$$ \frac1B \sum\limits^B_{i=1} [ 
    2   \la {f^*_{t_i}(x^\theta_{t_i,i})} - f_{\psi, t_i}(x^\theta_{t_i, i}), f^*_{t_i}(x^\theta_{t_i, i})  - s^{\theta}_{t_i, i}\ra  -  2  \cdot \alpha_{\text{SiD}}\|f^*_{t_i}(x^\theta_{t_i, i})  -  f_{\psi, t_i}(x^\theta_{t_i, i})\|^2];$$
    
\ENDFOR 
\end{algorithmic}
\end{algorithm}

\paragraph{Hyperparameters.} The only new hyperparameter introduced by our IDU method is the scale factor $\rho \in [0,1)$ which controls the strength of forgetting applied to the selected data. The theoretically justified value of $\rho = \pi$ can be approximately derived from the data split \eqref{eq: data split} as the proportion of the forget data within the overall dataset. Nevertheless, we still recommend trying other values of $\rho$ in practice to find the best trade-off between forgetting rate and retention quality. We demonstrate this trade-off for FM and SiD in Table~\ref{tab:cifar10_rho_ablation}. For the backbone-specific distillation scale, we use $\alpha_{\mathrm{SiD}}=0.5$ for FM and $\alpha_{\mathrm{SiD}}=1.2$ for SiD, following the corresponding original distillation recipes \citep{zhou2024score, kornilov2026universal}. {Further optimization details are provided in Appendix~\ref{app:optimization_hyperparameters}.}

\section{Experiments}
\label{sec:experiments}

\paragraph{Experimental setup.}

We evaluate IDU with flow matching (FM) \citep{tong2023improving} and the EDM-VP score-based backbone \citep{karras2024analyzing} used by Score identity Distillation (SiD) \citep{zhou2024score}, each on MNIST and CIFAR-10. The main experiments forget digits ${3,7}$ or CIFAR-10 classes ${1,9}$ (automobile and truck); the latter pair tests jointly removing related modes.
Class membership makes forgetting measurable, but IDU receives only forget samples, not labels or prompts. From each frozen full-data teacher, we run ordinary distillation and IDU, with the former verifying that the same framework also recovers a standard one-step generator when no forget data are mixed in. 
IDU optimization accesses only the teacher and forget samples. Retained images and evaluation classifiers are never supplied to the training objective.
Architectures, samplers, and hyperparameters are detailed in Appendices~\ref{app:architectures}, \ref{app:teacher_sampling}, and~\ref{app:optimization_hyperparameters}.

\paragraph{Evaluation protocol and metrics.}

We compute FID against the full training set for full-data teachers and ordinary distillation, and against retained data for IDU and retraining (\emph{Retain FID}). A teacher trained on retained data and its distilled student serve as retraining baselines. We also report the \emph{forgotten-class generation rate (FGR)}: the percentage of 50,000 generated images assigned to a forgotten class by an off-the-shelf classifier. Classifier and FID protocols are in Appendices~\ref{app:evaluation_classifiers} and~\ref{app:fid_protocol}. Unless noted, values are means and standard deviations over five evaluations; marked CIFAR-10 score-based values follow \citep{zhou2024score}.

\begin{table*}[t]
\centering
\caption{Main FM/SiD results on MNIST and CIFAR-10. FID uses full-data references for Pretrain/Pure distillation and retained-data references for Retrain/IDU; FGR is reported per forgotten class. Values are mean $\pm$ standard deviation over five runs, except $\dagger$ values from \citep{zhou2024score}.}
\label{tab:main_forgetting_results}
\resizebox{\textwidth}{!}{%
\begin{tabular}{lcccc}
\toprule
& \multicolumn{2}{c}{MNIST}
& \multicolumn{2}{c}{CIFAR-10} \\
\cmidrule(lr){2-3}
\cmidrule(lr){4-5}
Mode
& FID $\downarrow$
& \shortstack{FGR (\%) $\downarrow$\\Class 3 / Class 7}
& FID $\downarrow$
& \shortstack{FGR (\%) $\downarrow$\\Class 1 / Class 9} \\
\midrule

\multicolumn{5}{l}{\textbf{FM}} \\

Pretrain
& $0.88 \pm 0.01$
& \begin{tabular}[c]{@{}c@{}}
    $10.40 \pm 0.12$ \\
    $10.10 \pm 0.26$
  \end{tabular}
& $3.66 \pm 0.03$
& \begin{tabular}[c]{@{}c@{}}
    $12.50 \pm 0.08$ \\
    $11.18 \pm 0.12$
  \end{tabular} \\

Pure distillation
& $3.23 \pm 0.03$
& \begin{tabular}[c]{@{}c@{}}
    $10.50 \pm 0.15$ \\
    $10.53 \pm 0.09$
  \end{tabular}
& $4.35 \pm 0.05$
& \begin{tabular}[c]{@{}c@{}}
    $7.58 \pm 0.09$ \\
    $8.75 \pm 0.21$
  \end{tabular} \\

\rowcolor{gray!20}
{\bfseries Forgotten classes}
& \multicolumn{2}{c}{{\large $\{3,7\}$}}
& \multicolumn{2}{c}{{\large $\{1,9\}$}} \\

Retrain teacher
& $0.340 \pm 0.001$
& \textemdash
& $3.93 \pm 0.02$
& \textemdash \\

Retrain distillation
    & $5.72 \pm 0.12$
& \textemdash
    & $5.12 \pm 0.05$
& \textemdash \\

\midrule
\shortstack[l]{
IDU \\
{\scriptsize $(\rho_{\mathrm{MNIST}}=0.4)$} \\
{\scriptsize $(\rho_{\mathrm{CIFAR\text{-}10}}=0.6)$}
}
& $3.57 \pm 0.03$
& \begin{tabular}[c]{@{}c@{}}
    $0.16 \pm 0.01$ \\
    $0.16 \pm 0.01$
  \end{tabular}
& $5.81 \pm 0.05$
& \begin{tabular}[c]{@{}c@{}}
    $0.56 \pm 0.05$ \\
    $0.39 \pm 0.03$
  \end{tabular} \\

\midrule
\multicolumn{5}{l}{\textbf{SiD}} \\

Pretrain
& $1.35 \pm 0.02$
& \begin{tabular}[c]{@{}c@{}}
    $10.99 \pm 0.14$ \\
    $9.65 \pm 0.22$
  \end{tabular}
& $1.97^{\dagger}$
& \begin{tabular}[c]{@{}c@{}}
    $11.13 \pm 0.12$ \\
    $10.04 \pm 0.15$
  \end{tabular} \\

Pure distillation
& $1.12 \pm 0.01$
& \begin{tabular}[c]{@{}c@{}}
    $10.51 \pm 0.12$ \\
    $10.07 \pm 0.24$
  \end{tabular}
& $1.92 \pm 0.02\,^{\dagger}$
& \begin{tabular}[c]{@{}c@{}}
    $10.11 \pm 0.14$ \\
    $10.66 \pm 0.14$
  \end{tabular} \\

\rowcolor{gray!20}
{\bfseries Forgotten classes}
& \multicolumn{2}{c}{{\large $\{3,7\}$}}
& \multicolumn{2}{c}{{\large $\{1,9\}$}} \\

Retrain teacher
& $0.85 \pm 0.02$
& \textemdash
& $2.26 \pm 0.02$
& \textemdash \\

Retrain distillation
& $1.20 \pm 0.02$
& \textemdash
& $2.82 \pm 0.02$
& \textemdash \\

\midrule
\shortstack[l]{
IDU \\
{\scriptsize $(\rho=0.2)$}
}
& $1.65 \pm 0.02$
& \begin{tabular}[c]{@{}c@{}}
    $0.63 \pm 0.03$ \\
    $0.32 \pm 0.02$
  \end{tabular}
& $3.15 \pm 0.03$
& \begin{tabular}[c]{@{}c@{}}
    $1.11 \pm 0.06$ \\
    $1.25 \pm 0.06$
  \end{tabular} \\

\bottomrule
\end{tabular}%
}
\end{table*}

\paragraph{Main results.}
Table~\ref{tab:main_forgetting_results} shows that IDU suppresses both forgotten modes under all four dataset--backbone combinations. In the FM experiments, every forgotten-class FGR is at most $0.56\%$; in the SiD experiments, it is at most $1.25\%$. IDU also retains generation quality close to the corresponding retrain-distillation oracle in three settings and improves on that reference for FM on MNIST, indicating no substantial Retain FID degradation relative to target-matched retraining.
The FM/MNIST oracle comparison warrants caution owing to sensitivity of the retained-only retraining baseline; see Appendix~\ref{app:mnist_fm_details}. Visual results with Teacher--IDU sample grids appear in Appendix~\ref{app:visual_results}. \underline{Fine-tuning} experiments on the purely distilled generators achieve similar metrics; see Appendix \ref{app: finetune}.

\paragraph{Robustness across forget classes.}
We also evaluate single-class forgetting for all MNIST classes with FM and all CIFAR-10 classes with FM and SiD (Table~\ref{tab:single_class_robustness_combined}). Across these ten tasks per setting, mean Retain FID/FGR is $4.54/0.16\%$ for FM–MNIST, $7.73/0.84\%$ for FM–CIFAR-10, and $3.61/0.75\%$ for SiD–CIFAR-10. The per-class results show that low FGR is not confined to the pairs used in the main experiments.

\begin{table*}[t]
\centering
\caption{Single-class robustness of IDU with FM on MNIST and CIFAR-10 and with SiD on CIFAR-10. The final row averages the per-class means across the ten forget-class experiments.}
\label{tab:single_class_robustness_combined}
\small
\setlength{\tabcolsep}{6pt}
\begin{tabular}{c cc|cc|cc}
\toprule
& \multicolumn{2}{c|}{FM MNIST}
& \multicolumn{2}{c|}{FM CIFAR-10}
& \multicolumn{2}{c}{SiD CIFAR-10} \\
\cmidrule(lr){2-3}
\cmidrule(lr){4-5}
\cmidrule(lr){6-7}
\shortstack{Forgotten\\Class}
& Retain FID $\downarrow$ & FGR (\%) $\downarrow$
& Retain FID $\downarrow$ & FGR (\%) $\downarrow$
& Retain FID $\downarrow$ & FGR (\%) $\downarrow$ \\
\midrule
0
& $5.05 \pm 0.03$ & $0.10 \pm 0.01$
& $6.52 \pm 0.06$ & $0.76 \pm 0.03$
& $3.54 \pm 0.02$ & $0.58 \pm 0.03$ \\

1
& $3.46 \pm 0.04$ & $0.03 \pm 0.01$
& $6.45 \pm 0.09$ & $1.06 \pm 0.07$
& $2.96 \pm 0.01$ & $0.61 \pm 0.03$ \\

2   
& $3.84 \pm 0.02$ & $0.12 \pm 0.01$
& $7.83 \pm 0.05$ & $0.73 \pm 0.04$
& $4.07 \pm 0.05$ & $0.87 \pm 0.04$ \\

3
& $5.23 \pm 0.03$ & $0.10 \pm 0.01$
& $7.82 \pm 0.08$ & $1.51 \pm 0.04$
& $4.10 \pm 0.02$ & $1.11 \pm 0.02$ \\

4
& $4.71 \pm 0.03$ & $0.13 \pm 0.02$
& $8.53 \pm 0.08$ & $0.82 \pm 0.04$
& $4.13 \pm 0.03$ & $0.78 \pm 0.05$ \\

5
& $4.37 \pm 0.06$ & $0.17 \pm 0.02$
& $9.70 \pm 0.05$ & $1.18 \pm 0.06$
& $4.21 \pm 0.03$ & $0.88 \pm 0.04$ \\

6
& $4.39 \pm 0.04$ & $0.07 \pm 0.01$
& $8.45 \pm 0.09$ & $0.23 \pm 0.02$
& $4.16 \pm 0.05$ & $0.53 \pm 0.03$ \\

7
& $5.17 \pm 0.04$ & $0.15 \pm 0.01$
& $5.82 \pm 0.06$ & $1.23 \pm 0.05$
& $2.87 \pm 0.02$ & $0.90 \pm 0.05$ \\

8
& $4.22 \pm 0.04$ & $0.24 \pm 0.03$
& $8.88 \pm 0.13$ & $0.39 \pm 0.02$
& $3.22 \pm 0.02$ & $0.54 \pm 0.05$ \\

9
& $4.93 \pm 0.05$ & $0.45 \pm 0.03$
& $7.26 \pm 0.05$ & $0.47 \pm 0.03$
& $2.86 \pm 0.02$ & $0.72 \pm 0.01$ \\
\midrule
\textbf{Mean}
& $4.54$ & $0.16$
& $7.73$ & $0.84$
& $3.61$ & $0.75$ \\
\bottomrule
\end{tabular}
\end{table*}

\paragraph{Effect of the forgetting weight.}
Table~\ref{tab:cifar10_rho_ablation} varies the forget-mixture weight $\rho$ on CIFAR-10 for both FM and SiD. For FM, $\rho=0.9$ diverges immediately. Among the stable runs, low $\rho$ preserves FID but leaves higher FGR, whereas $\rho=0.8$ improves forgetting at a substantial FID cost. We therefore use $\rho=0.6$ as the best empirical balance for FM on CIFAR-10.
The FM coefficient for MNIST, $\rho=0.4$, is selected by the same quality--forgetting criterion. For SiD, increasing $\rho$ from $0.2$ to $0.8$ progressively worsens Retain FID, whereas the two class-wise FGR values vary non-monotonically. At $\rho=0.9$, SiD does not diverge, but its FGRs ($12.05\%$ and $9.55\%$) approach those of the full-data teacher, while Retain FID rises to $10.36$. Thus, excessive $\rho$ can degrade fidelity without forgetting.

\begin{table}[t]
\centering
\caption{Effect of the forget-mixture weight $\rho$ on CIFAR-10 for FM- and SiD-based IDU when jointly forgetting automobile (class 1) and truck (class 9). The selected configuration for each setup is bold; dashes denote settings not evaluated with SiD. FM training at $\rho=0.9$ diverges immediately.}
\label{tab:cifar10_rho_ablation}
\small
\setlength{\tabcolsep}{2pt}
\begin{tabular}{cccc@{\hspace{8pt}}cccc}
\toprule
\multicolumn{4}{c}{\textbf{Flow Matching}} & \multicolumn{4}{c}{\textbf{SiD}} \\
\cmidrule(lr){1-4}\cmidrule(lr){5-8}
$\rho$ & FID $\downarrow$ & FGR 1 (\%) $\downarrow$ & FGR 9 (\%) $\downarrow$
& $\rho$ & FID $\downarrow$ & FGR 1 (\%) $\downarrow$ & FGR 9 (\%) $\downarrow$ \\
\midrule
$0.05$ & $6.72 \pm 0.07$ & $3.04 \pm 0.08$ & $5.02 \pm 0.08$ & \multicolumn{4}{c}{\textemdash} \\
$0.1$ & $6.21 \pm 0.05$ & $3.75 \pm 0.09$ & $3.80 \pm 0.08$ & \multicolumn{4}{c}{\textemdash} \\
$0.2$ & $5.64 \pm 0.04$ & $3.35 \pm 0.03$ & $3.04 \pm 0.10$
& \textbf{$0.2$} & $\mathbf{3.15 \pm 0.03}$ & $\mathbf{1.11 \pm 0.06}$ & $\mathbf{1.25 \pm 0.06}$ \\
$0.4$ & $5.87 \pm 0.07$ & $0.62 \pm 0.06$ & $0.48 \pm 0.02$
& $0.4$ & $3.53 \pm 0.04$ & $0.74 \pm 0.05$ & $0.60 \pm 0.01$ \\
\textbf{$0.6$} & $\mathbf{5.81 \pm 0.05}$ & $\mathbf{0.56 \pm 0.05}$ & $\mathbf{0.39 \pm 0.03}$
& $0.6$ & $4.68 \pm 0.06$ & $1.16 \pm 0.05$ & $0.76 \pm 0.02$ \\
$0.8$ & $8.59 \pm 0.11$ & $0.38 \pm 0.03$ & $0.32 \pm 0.03$
& $0.8$ & $6.29 \pm 0.07$ & $0.81 \pm 0.05$ & $0.58 \pm 0.02$ \\
$0.9$ & \multicolumn{3}{c}{Diverged}
& $0.9$ & $10.36 \pm 0.10$ & $12.05 \pm 0.18$ & $9.55 \pm 0.12$ \\
\bottomrule
\end{tabular}
\end{table}

\section{Discussion and comparison}\label{sec:discussion}
\paragraph{How does our method work?} Our IDU leverages the teacher as guidance to steer generations away from unwanted data. Specifically, the IDU's generator loss  is a tractable form of the squared norm of the difference between the teacher $f^*$ and the fake model $f^{\mathrm{mix}}$ trained on the mixed forget and generated data $p_0^{\mathrm{mix}} := \rho \cdot p^F_0 +  (1-\rho)\cdot p^\theta_0$; see Appendix \ref{app:diff_proof}:
\begin{equation*}
 \mathcal{L}_{\text{IDU}}(f^{\mathrm{mix}},p_0^\theta)
=
\EE_{t,x_t\sim p_t^{\mathrm{mix}}}
[\left\|f_t^*(x_t)-f_t^{\mathrm{mix}}(x_t)\right\|^2],  \quad f^{\mathrm{mix}} := \argmin{_f}  \mathcal{L}_{\text{UM}}(f, p_0^{\mathrm{mix}}).
\end{equation*}
Since the supplied forget samples already cover the forget component of $p_0^{\mathrm{mix}}$, further forget samples generated by $G_\theta$ make the fake model depart from the teacher. Minimizing the generator loss counteracts this excess while also discouraging low-quality retained generations. %
Similar ideas are proposed for distillation acceleration in \citep{kornilov2026universal}. There, the authors utilize real data to push the generator toward the teacher faster, whereas we use  forget data to steer the model away.

\paragraph{Training and evaluation requirements.} 
IDU training uses the frozen teacher and forget samples only; retained images and external classifiers enter neither its losses nor its gradient updates.
On public MNIST and CIFAR-10, retained-data FID and classifier-based FGR provide direct, reproducible measurements.
We use these metrics to select $\rho$; without the retained data, $\rho$  can be chosen from a wide range of values, starting from the forget-set proportion, and still avoid significant degradation. However, the ablations show that larger values do not necessarily improve forgetting and may degrade fidelity. If direct evaluation is critical, retained data can be selected from the teacher samples via classification or manual selection.

\paragraph{Data unlearning methods.}  Our IDU is suitable for both the unlearning of the multi-step teacher matching models (with additional distillation) and the unlearning of the pretrained generators.

We begin by comparing the \textit{unlearning of matching models}. ContinualFlow \citep{simone2025continualflow} uses an energy function to guide a teacher flow model away from forget data, but it is limited to flow matching and relies on an image classifier to calculate the energy.
Other methods, such as NegGrad, SA \citep{heng2023selective}, SalUn \citep{fan2024salun}, SISS \citep{alberti2025siss}, Retrack \citep{shi2026retrack}, MGSM  \citep{Jiang_2025_ICCV}, EraseDiff \citep{wu2025erasing}, and VDU \citep{panda2024variational}, unlearn teacher models using a combination of forget and remaining losses \eqref{eq: siss}. Although our IDU also optimizes a similar combination of losses within the fake model, our working principle is different. The forget and remaining losses in other methods are trained adversarially: the former degrades quality on the forget dataset, while the latter preserves it on the remaining one. This is why such methods often resort to constrained optimization or more stable losses. In our IDU, by contrast, the forget and generated data distributions are modeled jointly as a mixture. As a result, we avoid multi-objective optimization and train our losses coherently.
Moreover, distilled generators provide efficient one-step inference rather than the multi-step sampling for the teacher.

\textit{For pretrained generators}, UOT-Unlearn \citep{choi2026uotunlearn} employs a cost function from unbalanced optimal transport to guide the unlearning. This guidance requires an additional feature extractor and manual cost-function tuning instead of a teacher model. Our IDU uses more complex teacher guidance but yields better forgetting and retaining metrics. As the UOT-Unlearn code is unavailable, we compare against their best unlearned CTM model with the SiD architecture and similar initial FID of 1.73. For CIFAR-10 unlearning classes 1, 6, and 8, their Retain FIDs ($\downarrow$) are (9.90, 5.11, 5.88) versus our (2.96, 4.16, 3.22), and their FGRs (\%, $\downarrow$) are around (2, 0.9, 1.5) versus our (0.61, 0.53, 0.54).
Many of the above matching model unlearning methods can be generalized to one-step consistency models. These models enforce self-consistency along the generation trajectory by mapping any point directly to the start. Thus, they can optimize different losses on the forget and remaining data.  For other one-step model types that do not take data samples as input, such as GANs, OT, and distillation, this generalization does not work. Moreover, such methods lag significantly behind in terms of retaining and forgetting (see Table 2 in \citep{choi2026uotunlearn}). In contrast, our IDU does not require self-consistency or a particular generator model type.

\paragraph{Class unlearning methods.} The most relevant SFD approach \citep{chen2025sfd} also performs unlearning during distillation but operates in the class unlearning setup. This method focuses on conditional models, where the data to erase is prompted through the input labels rather than through given samples. In contrast, our IDU can erase any part of training data, offering more flexible forgetting opportunities. The working mechanisms also differ dramatically. SFD modifies the generator loss \eqref{eq: Sid forget loss},  writing safe teacher information into the generator's unwanted classes. We modify the fake model loss to make it remember the mixture of generated and forget data and then compare it with the teacher's correct one, penalizing the generator for reproducing unwanted samples.
In the class-defined experiments reported in Tables~\ref{tab:main_forgetting_results}, \ref{tab:single_class_robustness_combined}, and \ref{tab:cifar10_rho_ablation}, the generator is never prompted with a class label: class annotations are used only to construct the forget subset and to evaluate FGR. Nevertheless, under the same conditions, our IDU achieves comparable results. In the CIFAR-10 class-0 forgetting experiment, SFD attains a Retain FID ($\downarrow$) of around 3.1 and an FGR (\%, $\downarrow$) of 0.36 (Figure 5 \citep{chen2025sfd}), versus 3.54 and 0.58 for our method (Table \ref{tab:single_class_robustness_combined}).

\paragraph{Optimization stability across backbones.}
In extended FM runs on MNIST and CIFAR-10, IDU often suppresses the target classes early, but their generation frequency can rise again after roughly 20k generator updates; the onset depends on $\rho$. FM therefore requires joint selection of $\rho$ and checkpoint. We did not observe this reversal over the evaluated SiD training horizon. This appears to be a limitation of the current FM instantiation, not of the IDU objective; larger, more stable FM architectures may allow $\rho$ to approach the forget-set proportion, as it does for SiD, where $\rho=0.2$ matches two forgotten classes out of ten.

\subsection*{AI use statement}
Generative AI tools were used solely to improve grammar, clarity, concision, and academic style during manuscript preparation. They were not used to formulate the research problem, develop the method, design or execute experiments, generate or analyze results, or determine the scientific claims and conclusions. All AI-assisted edits were reviewed and verified by the authors, who take full responsibility for the final content of this work.
\newpage
\bibliography{iclr2027_conference}
\bibliographystyle{iclr2027_conference}
\newpage
\appendix
\tableofcontents
\newpage
\section{{Proofs and related methods}}

\subsection{{Proof of Theorem~\ref{thm:idu_forgetting}}}
\label{app:idu_proof}
{

First, we need the formal convergence guarantees for the inverse distillation scheme \eqref{eq: inv general form 2}. 
\begin{lemma}[\textbf{Inverse distillation scheme's optimum \citep{kornilov2026universal}}] \label{lem: UID opt}
    The inverse scheme \eqref{eq: inv general form 2} with the teacher $f^* = \argmin_f \mathcal{L}_{\mathrm{UM}}(f, p^{*}_0)$ always attains its optimum $0$ when and only when the teacher data is retrieved, i.e., $p_0 = p_0^*$.
\end{lemma}

\begin{proofcustom}
According to Lemma \ref{lem: UID opt}, after we optimize the inverse distillation scheme \eqref{eq: inv general form 2} over distribution~$p_0$, we get the teacher data $p_0^*$ distilled into this optimized distribution, i.e., $p_0 = p_0^*$.
Since we parametrize the optimized distribution as a mixture of the generated and forget data $p_0 =  \rho\, p^{F}_0 +(1-\rho)\, p^{\theta}_0$ with $\rho = \pi$, then at the optimal parameters $\theta_{\text{opt}}$, we get:
\begin{eqnarray}
 p_0 = \rho\, p^{F}_0 +(1-\rho)\, p^{\theta_{\text{opt}}}_0 = \pi\, p^{F}_0 +(1-\pi)\, p^{\theta_{\text{opt}}}_0 = p_0^*.   \label{eq: split for opt params}
\end{eqnarray}
 Finally, considering the structure of the teacher data  \eqref{eq: data split}, we conclude that:
 \begin{eqnarray}
     \pi\, p^F_0 + (1-\pi)\, p^{\theta_{\text{opt}}}_0 \overset{\eqref{eq: split for opt params}}{=} p_0^* \overset{\eqref{eq: data split}}{=} \pi \, p^{F}_0 + (1-\pi)\, p^{R}_0   \quad \Rightarrow \quad p^{\theta_{\text{opt}}}_0 = p^{R}_0. \notag
  \end{eqnarray}
\end{proofcustom}

\subsection{The minimized distance}\label{app:diff_proof}

First, we use the formula of the IDU loss \eqref{eq: 4 losses} with the optimal fake model $f^{\mathrm{mix}} := \argmin{_f}  \mathcal{L}_{\text{UM}}(f, p_0^{\mathrm{mix}})$ on the current mixed data $p_0^{\mathrm{mix}} := \rho \cdot p^F_0 +  (1-\rho)\cdot p^\theta_0$:
\begin{equation*}
 \mathcal{L}_{\text{IDU}}(f^{\mathrm{mix}},p_0^\theta)
= \mathcal L_{\UM}(f^*,p^{\mix}_0)
-
\mathcal L_{\UM}(f^{\mathrm{mix}},p^{\mix}_0).
\end{equation*}

Following the structure of UM loss \eqref{eq: flow loss} on the mixed data, we get:
\begin{eqnarray*}
\mathcal L_{\UM}(f,p_0^{\mix})
&=&
\EE_{t,\,x_0\sim p_0^{\mix},\,x_t\sim p_t^{\mix}(\cdot\mid x_0)}
\left[
\left\|f_t(x_t)-f_t^{\mix}(x_t\mid x_0)\right\|^2
\right]\notag \\
&=&
\EE_{t,\,x_t\sim p_t^{\mix},\,x_0\sim p_0^{\mix}(\cdot\mid x_t)}
\left[
\left\|f_t(x_t)-f_t^{\mix}(x_t\mid x_0)\right\|^2
\right]\notag \\
&=&
\EE_{t,\,x_t\sim p_t^{\mix}}
\left[
\left\|f_t(x_t)-\EE_{x_0\sim p_0^{\mix}(\cdot\mid x_t)}[f_t^{\mix}(x_t\mid x_0)]\right\|^2
\right] + C(p^\mix_0),\notag \\
C(p^\mix_0) &=& \EE_{t,\,x_t\sim p_t^{\mix}}
\left[\EE_{x_0\sim p_0^{\mix}(\cdot\mid x_t)}\left[\left\|f_t^{\mix}(x_t\mid x_0)\right\|^2\right] \right]\notag \\
 &-&  \EE_{t,\,x_t\sim p_t^{\mix}}
\left[\left\|\EE_{x_0\sim p_0^{\mix}(\cdot\mid x_t)}\left[f_t^{\mix}(x_t\mid x_0)\right]\right\|^2
\right], \notag
\end{eqnarray*}
where $C(p^\mix_0)$ is the  bias--variance decomposition term independent of $f$. For fixed \(t,x_t\), the UM loss is minimized by the conditional mean:
\begin{eqnarray*}
f_t^{{\mix}}(x_t)
&=&
\EE_{x_0\sim p_0^{\mix}(\cdot\mid x_t)}[f_t^{\mix}(x_t\mid x_0)] = \argmin{_f}  \mathcal{L}_{\text{UM}}(f, p_0^{\mathrm{mix}}), \notag \\
\mathcal L_{\UM}(f,p_0^{\mix}) 
&=&
\EE_{t,\,x_t\sim p_t^{\mix}}
\left[
\left\|f_t(x_t)-f_t^{\mix}(x_t)\right\|^2
\right] + C(p^\mix_0).
\end{eqnarray*}

Thus, for the IDU loss with the optimal fake model $f^{\mix}$, we have:
\begin{eqnarray*}
\mathcal L_{\IDU}(f^{\mix},p_0^\theta)
&=&
\mathcal L_{\UM}(f^*,p_0^{\mix})
-
\mathcal L_{\UM}(f^{\mix},p_0^{\mix})
\\
&=&
\EE_{t,\,x_t\sim p_t^{\mix}}
\left[
\left\|f^*_t(x_t)-f_t^{\mix}(x_t)\right\|^2
\right] + C(p^\mix_0) \notag \\
&-&  \EE_{t,\,x_t\sim p_t^{\mix}}
\left[
\left\|f^\mix_t(x_t)-f_t^{\mix}(x_t)\right\|^2
\right] - C(p^\mix_0) \notag \\
&=& \EE_{t,\,x_t\sim p_t^{\mix}}
\left[
\left\|f^*_t(x_t)-f_t^{\mix}(x_t)\right\|^2
\right].
\end{eqnarray*}

\subsection{{SFD's details}}
\label{app:sfd_losses}
{
SFD \citep{chen2025sfd} distills a conditional teacher model into a one-step generator in parallel with class forgetting. It splits the generator loss from the inverse distillation scheme \eqref{eq: inv general form} into the forget and remaining losses as in \eqref{eq: siss}: in the forget loss, it aligns the conditional scores of the generator's forget classes $c_F$ with the teacher's scores for the safe classes $c_S$; in the remaining loss, it leaves other classes $c_R$ unswapped. The method also modifies these losses for better convergence, following the SiD framework \citep{zhou2024score}:
\begin{eqnarray}
    \mathcal{L}_{\text{SFD-gen-remain}}(p^\theta_0) &=& \EE_{t, x^\theta_0 \sim p^\theta_0(\cdot|c_R), x^\theta_t \sim p^\theta_t (\cdot| x^\theta_0, c_R)}  [ -  2  \cdot \alpha_{\text{SiD}}\|f^*_t(x^\theta_t|c_R)  - f_t(x^\theta_t|c_R)\|^2 \notag \\
    &+& 2   \la f^*_t(x^\theta_t|c_R) - f_t(x^\theta_t|c_R), f^*_t(x^\theta_t|c_R)  - f_t^{\theta}(x^\theta_t|x^\theta_0, c_R)\ra],\label{eq: Sid remain loss} \\
    \mathcal{L}_{\text{SFD-gen-forget}}(p^\theta_0) &=& \EE_{t, x^\theta_0 \sim p^\theta_0(\cdot|c_F), x^\theta_t \sim p^\theta_t(\cdot| x^\theta_0, c_F)}  [ -  2  \cdot \alpha_{\text{SiD}}\|f^*_t(x^\theta_t|c_S)  - f_t(x^\theta_t|c_F)\|^2 \notag \\
    &+& 2  \la f^*_t(x^\theta_t|c_S) - f_t(x^\theta_t|c_F), f^*_t(x^\theta_t|c_S)  - f_t^\theta(x^\theta_t|x^\theta_0, c_F)\ra],\label{eq: Sid forget loss}
\end{eqnarray}
where $\alpha_{\text{SiD}}$ is an arbitrary parameter, usually taken from the range $[0.5, 1.2]$. The loss for the fake model remains the same for all classes.
}

\newpage
\section{Experimental details}
\label{app:experimental_details}

\subsection{Architectures and teacher checkpoints}
\label{app:architectures}
For CIFAR-10, the FM setup uses the TorchCFM U-Net architecture from the public 400k-step Independent Conditional Flow Matching (I-CFM) checkpoint\footnote{\url{https://github.com/atong01/conditional-flow-matching/tree/main/examples/images/cifar10}}, whereas the SiD setup uses the DDPM++ (SongUNet) architecture of the public unconditional EDM-VP teacher adopted by the official SiD implementation\footnote{\url{https://github.com/mingyuanzhou/SiD}}. Both CIFAR-10 teachers operate on $32\times32$ RGB images. The public checkpoints are \path{cfm_cifar10_weights_step_400000.pt} for FM and \path{edm-cifar10-32x32-uncond-vp.pkl} for SiD.

For MNIST, we preserve each architecture family but adapt it to $28\times28$ grayscale inputs. In the FM U-Net, we change the input and output channels from 3 to 1, reduce the resolution hierarchy from \texttt{[1,2,2,2]} to \texttt{[1,2,2]}, move self-attention from resolution 16 to 14, reduce the number of channels per attention head from 64 to 32. 
In the SiD DDPM++ model, we change only the image resolution from 32 to 28, the image channels from 3 to 1, and the attention resolution from 16 to 14.

\subsection{Teacher sampling}
\label{app:teacher_sampling}
For the CIFAR-10 FM teacher, we use the adaptive Dormand--Prince (Dopri5) ODE solver with relative and absolute tolerances of $10^{-5}$, following the original TorchCFM setup. For the MNIST FM teacher, we use a fixed-step Euler solver with 100 integration steps. For both the MNIST and CIFAR-10 SiD teachers, we use the deterministic 18-step EDM sampler with the second-order correction prescribed by the original EDM and SiD implementations. Every distilled generator produces a sample in one step.

\subsection{Evaluation classifiers}
\label{app:evaluation_classifiers}
To compute FGR, we classify 50,000 generated images and report the percentage assigned to each forgotten class. For MNIST, we use the public LeNet-5 checkpoint\footnote{\url{https://github.com/hrfang/LeNet5-code-examples}}; its repository reports $99.13\%$ validation accuracy and $98.94\%$ test accuracy. For CIFAR-10, we use the public ResNet-56 checkpoint\footnote{\url{https://github.com/chenyaofo/pytorch-cifar-models}}, for which the repository reports $94.37\%$ top-1 and $99.83\%$ top-5 accuracy. These classifiers are external measurement instruments: they are not part of IDU, are never queried by the training loop, and contribute no loss, gradient, feature representation, or conditioning signal.

\subsection{FID protocol and evaluation data}
\label{app:fid_protocol}
All reported FID values use \texttt{clean-fid} v0.1.35 \citep{Parmar_2022_CVPR} in \texttt{legacy\_tensorflow} mode and 50,000 generated images. Images are quantized before extracting 2048-dimensional TensorFlow-compatible Inception features; grayscale MNIST samples are replicated across three channels. We use all available real training images from the distribution targeted by each row: the full training set for the full-data teacher and pure distillation, and the corresponding training subset with the forgotten labels removed for IDU and retained-only retraining. Thus, the paired-class references contain 60,000 full or 47,604 retained MNIST images and 50,000 full or 40,000 retained CIFAR-10 images. Single-class experiments analogously use the complete training subset excluding the selected class. Test images are not used as FID references.

Under the same legacy feature pipeline, the only numerical convention that differs from the original SiD evaluator is covariance normalization: CleanFID uses the sample covariance denominator $N-1$, whereas the original SiD/StyleGAN implementation uses the population denominator $N$. At 50,000 samples this changes the covariance scale only by the factor $N/(N-1)\approx1.00002$. Minor implementation-level differences remain in covariance symmetrization and numerical stabilization. Teacher FID uses the multi-step samplers described above; every distilled or IDU generator is evaluated with one-step inference.

\subsection{Optimization hyperparameters}
\label{app:optimization_hyperparameters}
\paragraph{FM on MNIST.} Full-data and retained-only teachers use learning rate $10^{-4}$, total batch size 128, and 100k and 50k optimizer steps, respectively. The retained-only teacher required fewer iterations because it empirically converged faster, likely due to the smaller amount of training data. Pure distillation, retained-only distillation, and IDU use learning rate $10^{-4}$, total batch size 256, and training horizons of approximately 30,000 generator iterations. Teacher pretraining uses Adam with $\beta=(0.9,0.999)$; distillation and IDU use Adam with $\beta=(0,0.999)$ for both trainable networks. All MNIST FM optimizers use cosine learning-rate annealing over their respective horizons.

\paragraph{FM on CIFAR-10.} The 400k-step teacher recipe uses learning rate $2\times10^{-4}$, global batch size 128, 5,000 warm-up steps, and 400,000 optimizer updates. Pure and retained-only distillation use learning rate $3\times10^{-5}$, total batch size 256, and approximately 50,000 generator iterations; IDU uses the same learning rate and horizon with configured per-process batch size 256. These runs use 500 warm-up steps. Across FM distillation and IDU runs, we use gradient clipping at 1, generator EMA 0.999, and $\alpha_{\mathrm{SiD}}=0.5$; the CIFAR-10 teacher itself uses EMA 0.9999.

\paragraph{SiD experiments.} MNIST teacher training uses learning rate $10^{-4}$, global batch size 512, and a 64-million-image (64 Mimg) horizon. The CIFAR-10 teacher recipe uses learning rate $10^{-3}$, global batch size 512, and a 200 Mimg training horizon; the full-data result uses the public EDM-VP checkpoint, while retained-only teachers follow this recipe. On both datasets, IDU, pure, and retained-only distillation use learning rates $10^{-5}$ for the generator and fake score network, global batch size 512, and a 100 Mimg training horizon. We retain $\alpha_{\mathrm{SiD}}=1.2$, $t_{\max}=800$, and initial noise standard deviation 2.5 from the original SiD recipe \citep{zhou2024score}.

These values are maximum optimization horizons rather than a claim that later checkpoints are always preferable. We select the reported checkpoint by the lowest observed FID; for FM, selection additionally precedes the late forgetting reversal discussed in Section~\ref{sec:discussion}.

\subsection{Initialization}
Our IDU can be used for both distillation from scratch and fine-tuning of an already pretrained generator. The only difference between the setups, besides the hyperparameter values, is the initialization: for fine-tuning, we initialize $G_\theta$ from the pretrained generator; otherwise, we initialize it from the one-step teacher inference scheme.

\subsection{Fine-tuning experiments} \label{app: finetune}
We evaluate fine-tuning on CIFAR-10 by initializing from the corresponding pure-distillation checkpoint and continuing IDU optimization with learning rates $5\times10^{-6}$ for FM and $10^{-6}$ for SiD, retaining the forgetting weights from the main experiments: $\rho=0.6$ for FM and $\rho=0.2$ for SiD. For forgotten classes 1 and 9, FM fine-tuning yields Retain FID $5.17\pm0.08$ and FGRs $0.89\pm0.02\%$ and $0.42\pm0.03\%$, respectively. SiD fine-tuning yields Retain FID $3.12\pm0.06$ and FGRs $2.69\pm0.06\%$ and $2.60\pm0.04\%$; see Table \ref{tab:finetune_forgetting_results} for comparison. Both fine-tuned models reduce forgotten-class generation relative to pure distillation, but the trade-off differs by backbone: compared with IDU trained from scratch, FM improves Retain FID while slightly increasing FGR, whereas SiD maintains comparable Retain FID but higher FGR.

\begin{table*}[t]
\centering
\caption{Fine-tune FM/SiD results for the purely distilled generators on CIFAR-10. FID uses full-data references for Pretrain/Pure distillation and retained-data references for IDU and fine-tuning; FGR is reported per forgotten class. Values are mean $\pm$ standard deviation over five runs, except $\dagger$ values from \citep{zhou2024score}.}
\label{tab:finetune_forgetting_results}
\resizebox{0.85\textwidth}{!}{%
\begin{tabular}{lcccc}
\toprule
& \multicolumn{2}{c}{FM}
& \multicolumn{2}{c}{SiD} \\
\cmidrule(lr){2-3}
\cmidrule(lr){4-5}
Mode
& FID $\downarrow$
& \shortstack{FGR (\%) $\downarrow$\\Class 1 / Class 9}
& FID $\downarrow$
& \shortstack{FGR (\%) $\downarrow$\\Class 1 / Class 9} \\
\midrule

Pretrain
& $3.66 \pm 0.03$
& \begin{tabular}[c]{@{}c@{}}
    $12.50 \pm 0.08$ \\
    $11.18 \pm 0.12$
  \end{tabular}
& $1.97^{\dagger}$
& \begin{tabular}[c]{@{}c@{}}
    $11.13 \pm 0.12$ \\
    $10.04 \pm 0.15$
  \end{tabular} \\

Pure distillation
& $4.35 \pm 0.05$
& \begin{tabular}[c]{@{}c@{}}
    $7.58 \pm 0.09$ \\
    $8.75 \pm 0.21$
  \end{tabular}
& $1.92 \pm 0.02\,^{\dagger}$
& \begin{tabular}[c]{@{}c@{}}
    $10.11 \pm 0.14$ \\
    $10.66 \pm 0.14$
  \end{tabular} \\

\rowcolor{gray!20}
{\bfseries Forgotten classes}
& \multicolumn{2}{c}{{\large $\{1,9\}$}}
& \multicolumn{2}{c}{{\large $\{1,9\}$}} \\
\midrule
\shortstack[l]{IDU from scratch \\
{\scriptsize $(\rho_{\mathrm{CIFAR\text{-}10}}=0.6)$} \\
{\scriptsize $(\rho_{\mathrm{SiD}}=0.2)$}}
& $5.81 \pm 0.05$
& \begin{tabular}[c]{@{}c@{}}
    $0.56 \pm 0.05$ \\
    $0.39 \pm 0.03$
  \end{tabular}
& $3.15 \pm 0.03$
& \begin{tabular}[c]{@{}c@{}}
    $1.11 \pm 0.06$ \\
    $1.25 \pm 0.06$
  \end{tabular} \\
\midrule
\shortstack[l]{IDU Fine-tuning \\
{\scriptsize $(\rho_{\mathrm{CIFAR\text{-}10}}=0.6)$} \\
{\scriptsize $(\rho_{\mathrm{SiD}}=0.2)$}}
& $5.17 \pm 0.08$
& \begin{tabular}[c]{@{}c@{}}
    $0.89 \pm 0.02$ \\
    $0.42 \pm 0.03$
  \end{tabular}
& $3.12 \pm 0.06$
& \begin{tabular}[c]{@{}c@{}}
    $2.69 \pm 0.06$ \\
    $2.60 \pm 0.04$
  \end{tabular} \\

\bottomrule
\end{tabular}%
}
\end{table*}

\subsection{Code and checkpoint release}
Upon publication, we will release the complete source code, exact configurations, evaluation scripts, and checkpoints used to produce the main reported results. 

\subsection{MNIST FM retraining and sensitivity to $\rho$}
\label{app:mnist_fm_details}
The retained-only FM baseline on MNIST exhibits a distinct sensitivity. The exceptionally low Retain FID of the retrained teacher ($0.34$) may reflect strong overfitting to the smaller retained subset rather than uniformly better generalization. In its distilled counterpart, we consistently observed poor generation of digit 2, which raises Retain FID to $5.72$. This failure occurred despite using the same architecture and optimization settings as the stable full-data pretraining and distillation runs, suggesting sensitivity of the retraining baseline to the altered data distribution rather than an intentional hyperparameter disadvantage.

Table~\ref{tab:mnist_rho_ablation} reports the FM/MNIST results for different $\rho$. The main setting, $\rho=0.4$, gives the best observed Retain FID--FGR trade-off. Increasing $\rho$ to $0.6$ or above drives the target-class FGR values to zero, but also suppresses additional digits---most consistently 2 and 5, with 8 or 9 affected in some runs---and raises Retain FID above 10. Conversely, decreasing $\rho$ below $0.4$ progressively weakens forgetting of digits 3 and 7 and does not improve Retain FID over the main setting. This sensitivity appears specific to the low-dimensional MNIST setting and the FM architecture and training recipe used here, and motivates evaluation with larger and more stable backbones. In contrast, the more recent SiD distillation backbone remained stable at $\rho=0.2$, which matches the nominal fraction of two forgotten classes among ten, without the same collateral class suppression.

\begin{table*}[t]
\centering
\caption{Effect of the forget-mixture weight $\rho$ on FM-based IDU for MNIST when jointly forgetting digits 3 and 7. The main configuration is bold. The final column lists qualitatively suppressed digits in addition to the target digits 3 and 7.}
\label{tab:mnist_rho_ablation}
\small
\setlength{\tabcolsep}{8pt}
\begin{tabular}{ccccc}
\toprule
$\rho$ & Retain FID $\downarrow$ & FGR 3 (\%) $\downarrow$ & FGR 7 (\%) $\downarrow$ & Additional suppressed digits \\
\midrule
$0.05$ & $4.29 \pm 0.02$ & $7.07 \pm 0.06$ & $8.18 \pm 0.06$ & \textemdash \\
$0.1$  & $3.96 \pm 0.04$ & $3.87 \pm 0.09$ & $4.62 \pm 0.10$ & \textemdash \\
$0.2$  & $4.81 \pm 0.08$ & $2.22 \pm 0.07$ & $1.63 \pm 0.07$ & \textemdash \\
$\mathbf{0.4}$ & $\mathbf{3.57 \pm 0.03}$ & $\mathbf{0.16 \pm 0.01}$ & $\mathbf{0.16 \pm 0.01}$ & \textemdash \\
$0.6$  & $11.20 \pm 0.10$ & $0$ & $0$ & 2, 5, 8 \\
$0.8$  & $10.58 \pm 0.07$ & $0$ & $0$ & 2, 5 \\
$0.9$  & $10.35 \pm 0.04$ & $0$ & $0$ & 2, 5, 9 \\
\bottomrule
\end{tabular}
\end{table*}

\clearpage
\section{{Visual Results}}
\label{app:visual_results}

{We qualitatively compare samples from each multi-step teacher with samples from the corresponding one-step IDU generator. For MNIST, IDU is trained to forget digits 3 and 7; for CIFAR-10, it is trained to forget automobile (class 1) and truck (class 9). Across both FM and SiD, the teacher grids contain the target classes, whereas the IDU grids visibly suppress them while preserving samples from the retained classes. These finite grids are intended as qualitative illustrations; the corresponding 50,000-sample FGR measurements are reported in Table~\ref{tab:main_forgetting_results}.}

\begin{figure}[!htbp]
    \centering
    \begin{minipage}[t]{0.485\textwidth}
        \centering
        \includegraphics[width=\linewidth]{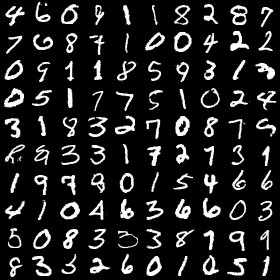}
        \par\smallskip{\small Teacher}
    \end{minipage}
    \hfill
    \begin{minipage}[t]{0.485\textwidth}
        \centering
        \includegraphics[width=\linewidth]{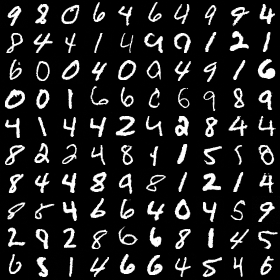}
        \par\smallskip{\small IDU}
    \end{minipage}
    \caption{{FM on MNIST: teacher and IDU samples when forgetting digits 3 and 7.}}
    \label{fig:visual_mnist_fm_full}
\end{figure}

\begin{figure}[!htbp]
    \centering
    \begin{minipage}[t]{0.485\textwidth}
        \centering
        \includegraphics[width=\linewidth]{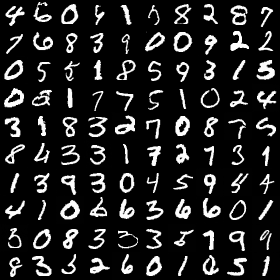}
        \par\smallskip{\small Teacher}
    \end{minipage}
    \hfill
    \begin{minipage}[t]{0.485\textwidth}
        \centering
        \includegraphics[width=\linewidth]{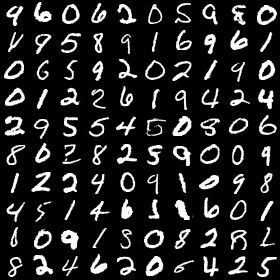}
        \par\smallskip{\small IDU}
    \end{minipage}
    \caption{{SiD on MNIST: teacher and IDU samples when forgetting digits 3 and 7.}}
    \label{fig:visual_mnist_sid_full}
\end{figure}

\begin{figure}[!htbp]
    \centering
    \begin{minipage}[t]{0.485\textwidth}
        \centering
        \includegraphics[width=\linewidth]{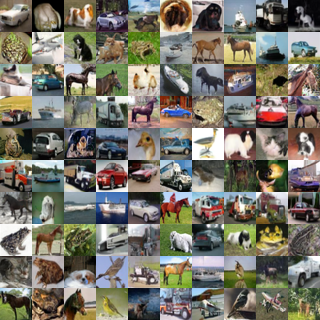}
        \par\smallskip{\small Teacher}
    \end{minipage}
    \hfill
    \begin{minipage}[t]{0.485\textwidth}
        \centering
        \includegraphics[width=\linewidth]{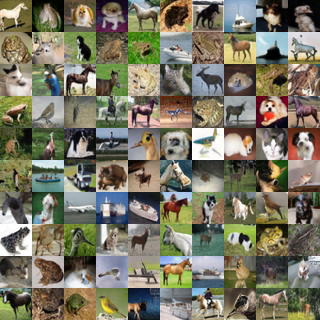}
        \par\smallskip{\small IDU}
    \end{minipage}
    \caption{{FM on CIFAR-10: teacher and IDU samples when forgetting automobile (class 1) and truck (class 9).}}
    \label{fig:visual_cifar10_fm_full}
\end{figure}

\begin{figure}[!htbp]
    \centering
    \begin{minipage}[t]{0.485\textwidth}
        \centering
        \includegraphics[width=\linewidth]{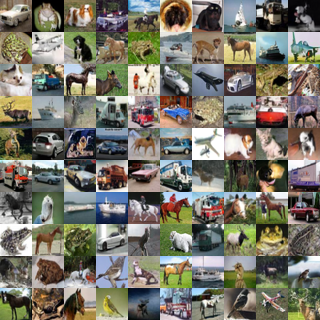}
        \par\smallskip{\small Teacher}
    \end{minipage}
    \hfill
    \begin{minipage}[t]{0.485\textwidth}
        \centering
        \includegraphics[width=\linewidth]{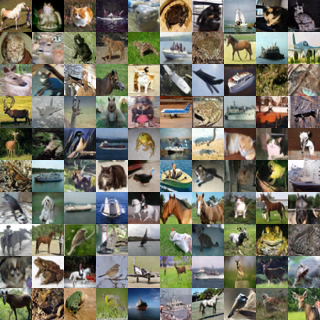}
        \par\smallskip{\small IDU}
    \end{minipage}
    \caption{{SiD on CIFAR-10: teacher and IDU samples when forgetting automobile (class 1) and truck (class 9).}}
    \label{fig:visual_cifar10_sid_full}
\end{figure}

\end{document}